\documentclass{article}
\pdfoutput=1 
\usepackage[utf8]{inputenc}
\usepackage{subcaption}
\usepackage{graphicx}
\usepackage{threeparttable}
\usepackage{amsmath, bm}
\usepackage[dvipsnames]{xcolor}
\usepackage{hyperref}
\usepackage{cleveref}
\newcommand\snr{SNR optimality}
\newcommand\infomax{the \emph{infomax} principle}
\newcommand\snrgain{SNR gain}
\newcommand\relu{ReLU}
\newcommand\sign{\text{sgn}}

\newcommand\myshade{90}
\colorlet{mylinkcolor}{NavyBlue}
\colorlet{mycitecolor}{Aquamarine}
\colorlet{myurlcolor}{Aquamarine}

\hypersetup{
  linkcolor  = mylinkcolor!\myshade!black,
  citecolor  = mycitecolor!\myshade!black,
  urlcolor   = myurlcolor!\myshade!black,
  colorlinks = true,
}

\usepackage[margin=2.6cm]{geometry}
\usepackage{libertine}
\usepackage{libertinust1math}
\usepackage{inconsolata}
\usepackage[T1]{fontenc}    
\usepackage[sf,bf]{titlesec}

\usepackage{parskip}
\renewcommand{\vec}[1]{\boldsymbol{#1}}

\usepackage[backend=bibtex,
style=numeric,
sorting=none,
isbn=false,
doi=false,
url=false,
]{biblatex}

\title{\textsf{\textbf{Think Global, Act Local:\\ Relating DNN generalisation and node-level SNR}}}
\author{
  Paul Norridge\\
  \texttt{\small paul.norridge@gmail.com}
  }

\begin{document}

\maketitle
\begin{abstract}
The reasons behind good DNN generalisation remain an open question. In this paper we explore the problem by looking at the Signal-to-Noise Ratio of nodes in the network. Starting from information theory principles, it is possible to derive an expression for the SNR of a DNN node output. Using this expression we construct figures-of-merit that quantify how well the weights of a node optimise SNR (or, equivalently, information rate). Applying these figures-of-merit, we give examples indicating that weight sets that promote good SNR performance also exhibit good generalisation. In addition, we are able to identify the qualities of weight sets that exhibit good SNR behaviour and hence promote good generalisation. This leads to a discussion of how these results relate to network training and regularisation. Finally, we identify some ways that these observations can be used in training design.
\end{abstract}

\section{Introduction}
Deep neural networks are being used ever more widely for machine learning applications and continue to demonstrate impressive results. However, there still remain aspects of neural network optimisation and performance that cannot be explained fully. In particular, the precise reason why some networks generalise better than others is not always clear \cite{Chiyuan_Zhang_et_al_2016}. \par

The aim of this paper is to highlight a contributor to DNN performance not typically considered in generalisation discussions: the robustness that the trained network has towards (random) noise and the weight configurations that enhance this. Of course, the effect of noise is primarily an information theory consideration. Following Linsker \cite{R._Linsker1988}, information theory approaches and, in particular, \infomax{}, are frequently applied to DNN analysis and training. These typically focus on mutual information between node inputs and outputs. Instead, in this paper we will consider a related quantity: the Signal-to-Noise Ratio (SNR) of each of the nodes in the network. Linsker derived in \cite{R._Linsker1988} an expression for information rate in terms of SNR.  We take this as our starting point. \par
 
As will be seen, working in terms of SNR has a number of benefits. In particular, we are able to assess in a straightforward way to what extent the weights associated with a node enhance the information rate. We derive two figures-of-merit that respectively (a) measure the deviation of weights from optimal SNR/rate maximisation (effectively assessing them against an \emph{infomax-like} criteria) and (b) measure how much a node enhances SNR compared to its inputs. Once we have these criteria, we are able to two further steps: First, we can test how the SNR properties of the network relate to the generalisation performance; we give we give examples indicating that weight sets that promote good SNR performance also exhibit good generalisation. Surprisingly, this relationship can be demonstrated even though the interactions between nodes in the network are complex and there is no a priori characterisation of what constitutes \emph{signal} as opposed to \emph{noise} on a node output. Second, we can identify the properties of node weights that maximise SNR and, hence, generalisation. \par   

Based on observations about the qualities of an SNR-optimising weight set, we discuss how SNR optimisation relates to other aspects of DNN training and why we expect common regularisation schemes to promote SNR-optimal weights.\par

Although it is beyond the scope of this paper to go into the details of applications, this is not intended to be a purely abstract study. Consequently, we conclude the discussion by highlighting ways that an SNR perspective can aid network training and use. \par

\section{Related Work}

The question of what ensures good generalisation of a DNN is a long-standing question, given renewed impetus by \cite{Chiyuan_Zhang_et_al_2016}.  The investigations into generalisation are extensive, so it is not possible to consider the full range; we focus on those relevant for our discussion. The results in \cite{Chiyuan_Zhang_et_al_2016} raised the primary question of how we align the generalisation ability of DNNs with the very large capacity. Alongside this were the more general uncertainties over how we explain generalisation and the role of regularisation. There have been a number of studies related to the generalisation/capacity question. One focus has been better bounds for classification and risk \cite{ExploringGeneralization, 2017arXiv171101530L, 2017arXiv171005468K}, with consideration of how generalisation and over-fitting interact with minima sharpness/flatness \cite{Sepp_Hochreiter_Jrgen_Schmidhuber1997, Nitish_Shirish_Keskar_et_al_2016}, robustness, weight norms and margins. Bounds have also been considered in stochastic neural networks and via PAC-Bayes analysis \cite{2018arXiv180405862Z, 2017arXiv170311008K}. In addition, there have been approaches have looking at learning dynamics \cite{2018arXiv180611379P} and Bayesian analysis of SGD \cite{2017arXiv171006451S}. A general assumption in all of these is that over-fitting is the key question, however there is also a suggestion that the significance of the capacity question is reduced by other factors, for example the `interpolation regime' coming from large networks \cite{2018arXiv181211118B} and the global minimum selection \cite{2019arXiv190603830A}. In this paper we focus on the more general question of explaining generalisation and the role of regularisation. Rather than directly addressing the question of capacity and over-fitting, our results come from looking at features of network weights that impact performance beyond straightforward fit to the data. However, even though we come from a different direction, it will be apparent that our discussion is relevant to some of the metrics listed above. \par

Another approach to characterising generalisation performance has been to look at information theory bounds. As with the work here, these use mutual information or information rate as a starting point. This may be done at network-level \cite{2018arXiv180409060Z}, or, more commonly, in the context of the learning algorithm rather than a trained network \cite{ChainingMutualInformation, 2019arXiv190104609B, Xu2017InformationtheoreticAO}. Again, the assumption behind the majority of these analyses is that minimisation of over-fitting is the dominant issue for generalisation.\par 

The use of mutual information in network training and analysis has a long history, starting with \infomax{} proposed by Linsker \cite{R._Linsker1988}. \emph{Infomax} has been used directly in DNN training both end-to-end \cite{2019arXiv190511786L} and for training of individual layers (for example, \cite{2018arXiv180703748V, hjelm2018learning}). Mutual information has also been used successfully in the ranked selection of features (see, for example \cite{Brown2009ANP}) and for data augmentation \cite{DiscreteRepresentations}. The most successful recent application is in the development of the \emph{Information Bottleneck Principle} \cite{2015arXiv150302406T, 2019arXiv190403743H, Elad2018TheEO}. This uses expressions for mutual information between layers both to characterise and as a basis for training. \par 

In order to optimise the applications of mutual information, there are a number of recent steps towards better estimates for mutual information \cite{2019arXiv190506922P, EntropyAndMutualInformation}. In contrast to these, the work here avoids more rigorous approaches in favour of estimating how well a given set of weights optimises this quantity.  \par  

\section{Groundwork}\label{sec.groundwork}

With obvious echoes of the \emph{infomax} principle, the underlying reasoning for considering node-level SNR and SNR-optimising weights can be summarised as:
\begin{enumerate}
\item{A DNN will perform best when the available (useful) information is exploited to maximum extent.}
\item{Optimal use of information by the entire network depends on maximising the information preservation of individual nodes.}
\item{With some assumptions, it is possible to derive a relationship between the output SNR of a node and the maximisation of the information preservation from inputs to output (characterised by the mutual information or the information rate at the output of the node). So, if we want to optimise the information flow in the network, we should pay attention to the SNR of the individual nodes.}
\item{It is possible to quantify how well a given set of weights optimises the SNR within the context of a given network and training set. Even though we cannot define a priori the \emph{signal} that a node should generate, we can construct formulae that allow the SNR quality of a post-training weight set to be assessed.}
\end{enumerate}

An alternative way to express this is to say that network performance will be better when noise sensitivity is low. And while many weight combinations can fit to a given data set, some weight sets are more sensitive to noise than others. \par

Here, we start with the node-level characterisation and then provide examples that support the relationship to network performance.\par

For all of the networks considered here, our building block is the usual node definition 
\begin{equation} 
g\left(\sum_jw_{ij}x_{j} + b_i\right) 
\end{equation}
with inputs $x_j$, weights $w_{ij}$, bias $b_i$ and activation function $g()$.\par

For the SNR assessment of this node, what is important is the weighted inputs
\begin{equation} 
y_i = \sum_jw_{ij}x_j
\end{equation}
Partitioning the inputs into signal and noise components, this becomes:
\begin{equation} 
y_i = \sum_jw_{ij}\left(s_j + n_j\right) 
\end{equation}
At this stage, we leave open the question of how we identify the two components.\par

Following the Shannon's derivations \cite{Shannon}, Linsker \cite{R._Linsker1988} has shown that if $y_i$ and $n_i$ are Gaussian, maximising the information rate at the output of a node is equivalent to maximisation of an SNR-like expression\footnote{A similar expression for SNR derived in a very different -- but mathematically equivalent -- context can be found in \cite{AntennaGT}.}. \par
For node $i$ in layer $m$, we express the information rate as 
\begin{equation}
R_i^{(m)}=\frac{1}{2} \ln\left(SNR_i^{(m)}\right)
\end{equation}
with 
\begin{equation}\label{eq:snr}
SNR_i^{(m)} =\frac{\text{var} \left(\sum_{j}w_{ij}s_{j} \right)}{ \text{var} \left( \sum_{j}w_{ij}n_{j} \right) } = \frac{\text{var} \left(\sum_{j}w_{ij}s_{j} \right)}{ \sum_{j}w_{ij}^2\text{var} \left( n_{j} \right) } 
\end{equation}
For the expansion of the noise variance, we have assumed that the noise components from different inputs are statistically independent. \par

Our intention is to generate a pragmatic characterisation of nodes in a network, so rather than consider the probability distributions of the quantities in a rigorous manner, we rely on the fact that nodes will typically have a large number of inputs and, consequently, equation (\ref{eq:snr}) is a good approximation. All variances (and, later, covariances) are calculated over a batch of samples.\par

Starting from equation (\ref{eq:snr}), let us look for weights that maximise the expression\footnote{It is interesting to note that we can view $l_2$ regularisation as an alternative route taken from the same starting point. It can be interpreted as a way to minimise the denominator of (\ref{eq:snr}) rather than maximising the total expression.}. Differentiating with respect to the $w_{ij}$, we find \par

\begin{equation}
w_{ij}= k_i .  \frac{\text{cov} \left( s_j, \sum w_{ij} s_i \right) }{\text{var} \left( n_j \right)}  
\end{equation}

Where  $ k_i $  is a constant independent of  \( j \)  and we have implicitly assumed that  $ \text{var} \left( n_{j} \right)\neq 0 $ .\par

To allow us to use this, we make two further pragmatic assumptions. 
The first is that the after convergence, the training process has identified an appropriate `signal' for this node and that signal dominates the node output, so that 
\begin{equation}
\text{cov} \left( s_i, \sum w_{ij} s_j \right) \approx \text{cov} \left( x_j, \sum w_{ij} x_i \right)
\end{equation}

The second is that the noise is Gaussian with identical variance for each non-zero input sample\footnote{Of course, this is hard to justify generically, but is assumed pragmatically. Further we make normalisation choices when implementing that makes this more likely. Not taking this approach complicates the computation significantly and tests suggest it only provides minimal gain.}.\par
With a ReLU activation on the input, noise will only contribute when the $x_i>0$. That is, the noise will have zero variance whenever $x_i<0$\footnote{We observe that when $x_i$ is close to zero the assumption that the noise contribution is either Gaussian or zero will be violated, but this effect is not considered to be significant enough to affect the analysis here.}. Combining this observation with the above assumption, we make the approximation
\begin{equation}
 \text{var} \left( n_{j} \right) \approx a_j ~ \text{var}(n)
\end{equation}

where $\text{var}(n)$ is the common noise variance of all (non-zero) samples arriving at node $i$ and $a_i$ is the rate of activations of node $j$ of layer $(m-1)$, calculated as  $a_i = p(x_i > 0)$.

Using these assumptions, we can update the expression for optimal weights to
\begin{equation}
w_{ij}^{opt} = k_i' .  \frac{\text{cov} \left( x_i, \sum w_{ij} x_j \right) }{ a_i }  
\end{equation}

We do not expect that nodes will generally meet this condition, but we would like to assess how close a given weight configuration is to `optimal'. We can measure this by treating $w_{ij}$ and $\frac{\text{cov} \left( x_i, \sum w_{ij} x_j \right) }{a_j}$ as vectors and calculating the inner product between them\footnote{At this point, we note that there is a close relationship between this condition and PCA. When $a_j = 1 \ \forall j$, the weight will be optimal if it is an eigenvector of the covariance matrix. See also the observation in \cite{R._Linsker1988}}. This leads to the expression

 \begin{equation}\label{eq:snropt}
     S_i =\frac{\langle\vec{w_i},\vec{w_i^{opt}}\rangle}{ \| \vec{w_{i}} \|  \| \vec{w_i^{opt}} \| }=\frac{ \sum _{i}w_{ij}c_{j}}{ \| w_{i} \|  \| \vec{c} \| }
 \end{equation}

where $\vec{c}$ is the vector with components
$c_{i}= \frac{\text{cov} \left( x_i,\sum w_{ij} x_j\right)}{a_i} 
$\ and $\vec{w_i}$ is the vector with components $w_{ij}$.\par
In the following we will refer to $S_i$ as the \emph{\snr{}} and it will act as the basic building block of our analysis.\par

The following steps will assert that this quantity can be used as a tool for network characterisation. It is surprising that the network-level performance can be so closely related to a characteristic of individual nodes, so we make the argument in stages. First we apply equation (\ref{eq:snropt}) to a simple example of a fully connected DNN to show how the node-level \snr{} varies with generalisation performance. This will give strong indications that SNR considerations are important, but we find that comparisons become harder when applying it to more complex examples. For this reason, we generate a second figure-of-merit, based on $S_i$, that is more useful generally. This refined figure-of-merit is applied to both fully connected and CNN examples.\par

Although we focus on the quantitative measures of \snr{}, we emphasise that the primary goal is to understand better what makes a network successful. That is, the aim is to demonstrate that DNN performance \emph{is} related to weight configurations that optimise SNR. This in turn demonstrates that information preservation is an important principle for DNN performance, even when it is not used as an explicit optimisation goal.\par

\section{\snr{} applied to MNIST}

In order to demonstrate that node-level ‘\snr{}’ is correlated with the overall performance of a DNN, we first look at a simple example. We repeatedly train a 3-layer, fully-connected network for classification on the MNIST dataset \cite{MNIST, lecun2010mnist}, using a variety of different training conditions. Then, using the test set, we compare the \snr{} for the different trained networks against the accuracy. \par

Even in a simple network such as this, considering the \snr{} for every node in the network would be unmanageable. To simplify interpretation, we instead look layer-by-layer and calculate the average \snr{} across each layer as a whole. In taking this approach we accept that there may be subtleties that are obscured, but we will see that in practice it generates useful results. \par 

\begin{figure}[!htbp]
     \begin{subfigure}[b]{0.32\linewidth}
       \includegraphics[width=\linewidth]{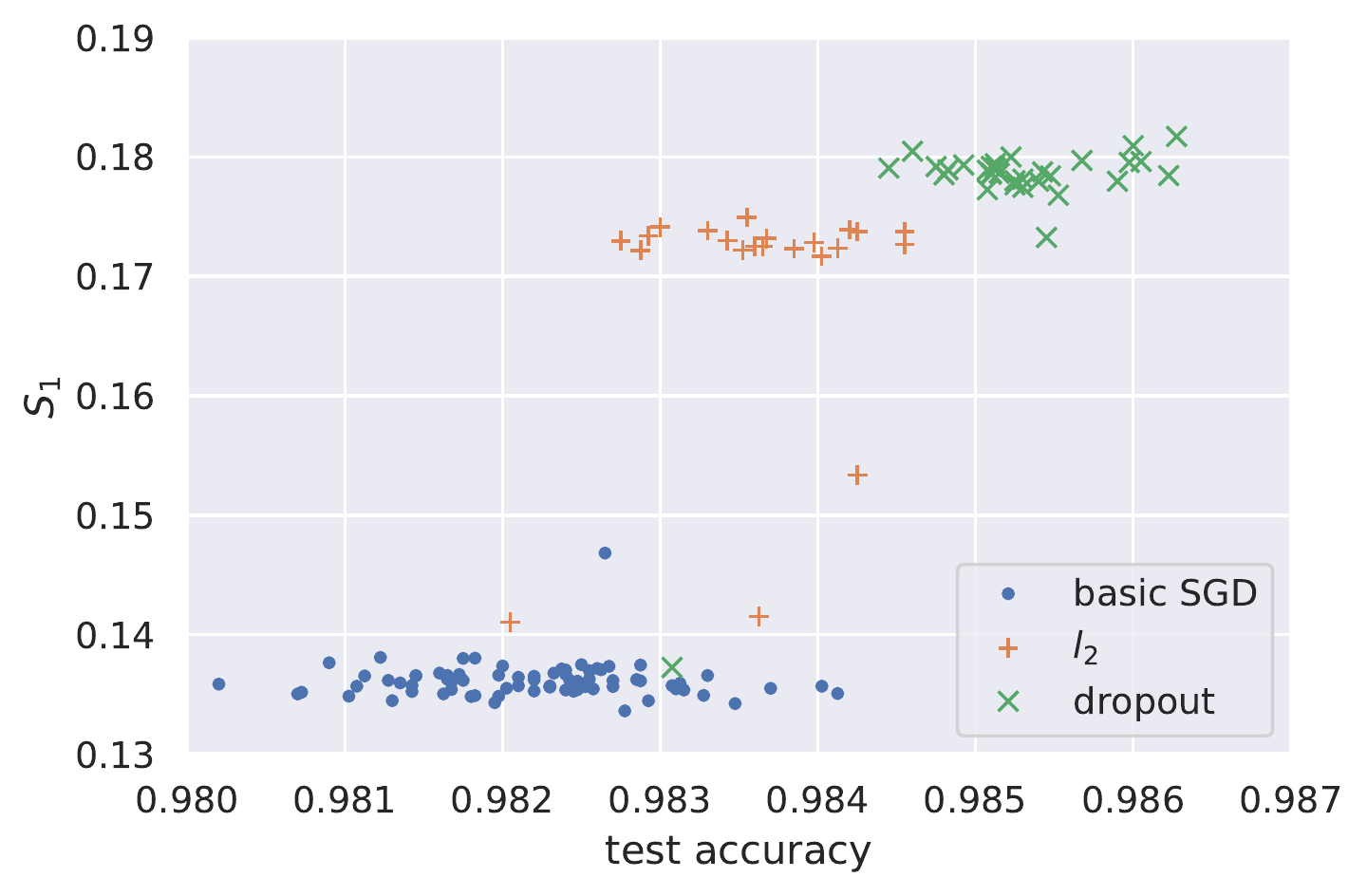}
       \caption{\snr{} for Layer 1}

     \end{subfigure}
     \hfill
     \begin{subfigure}[b]{0.32\linewidth}
       \includegraphics[width=\linewidth]{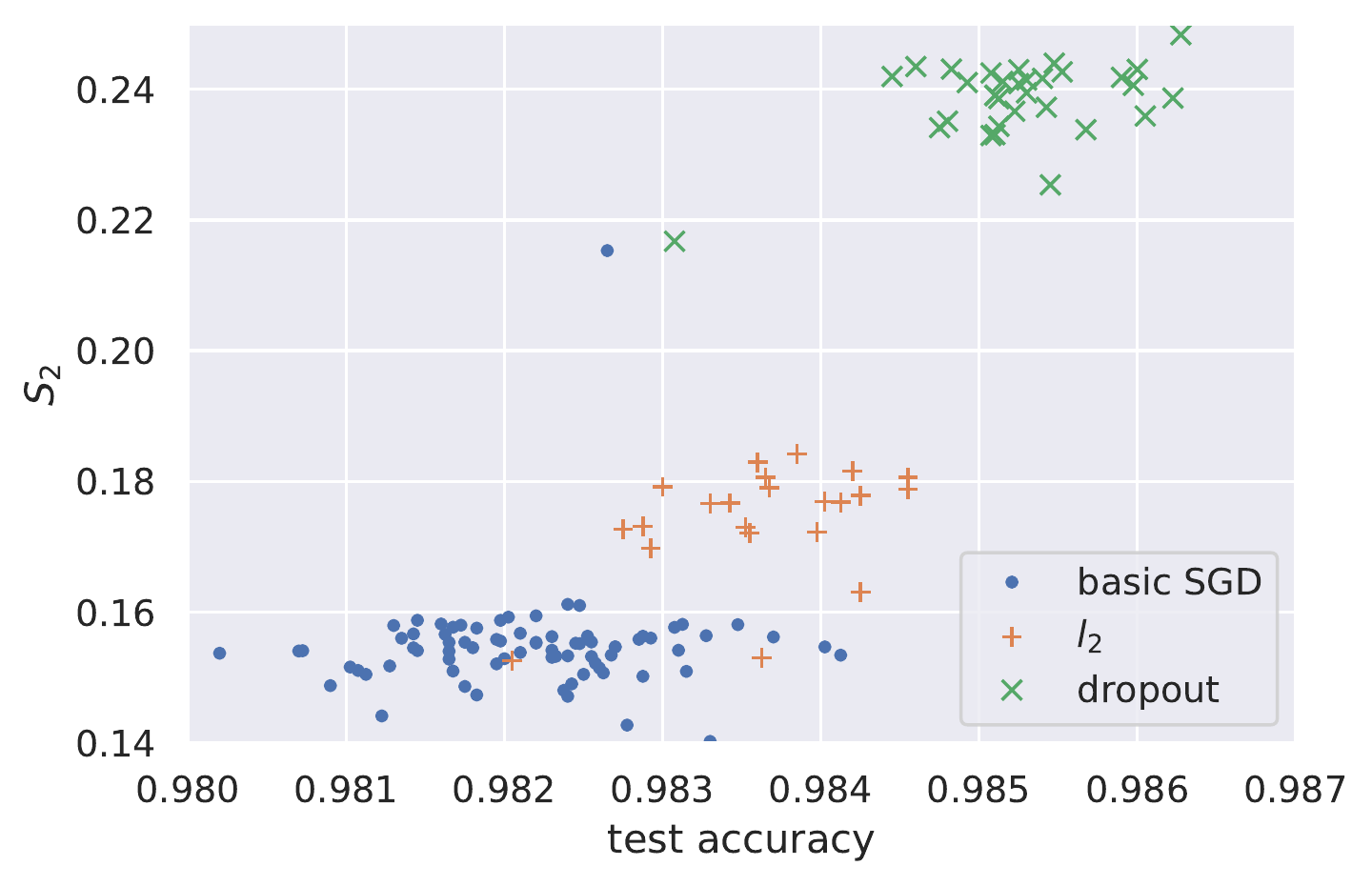}
       \caption{\snr{} for Layer 2}
     \end{subfigure}
    \hfill
     \begin{subfigure}[b]{0.32\linewidth}
       \includegraphics[width=\linewidth]{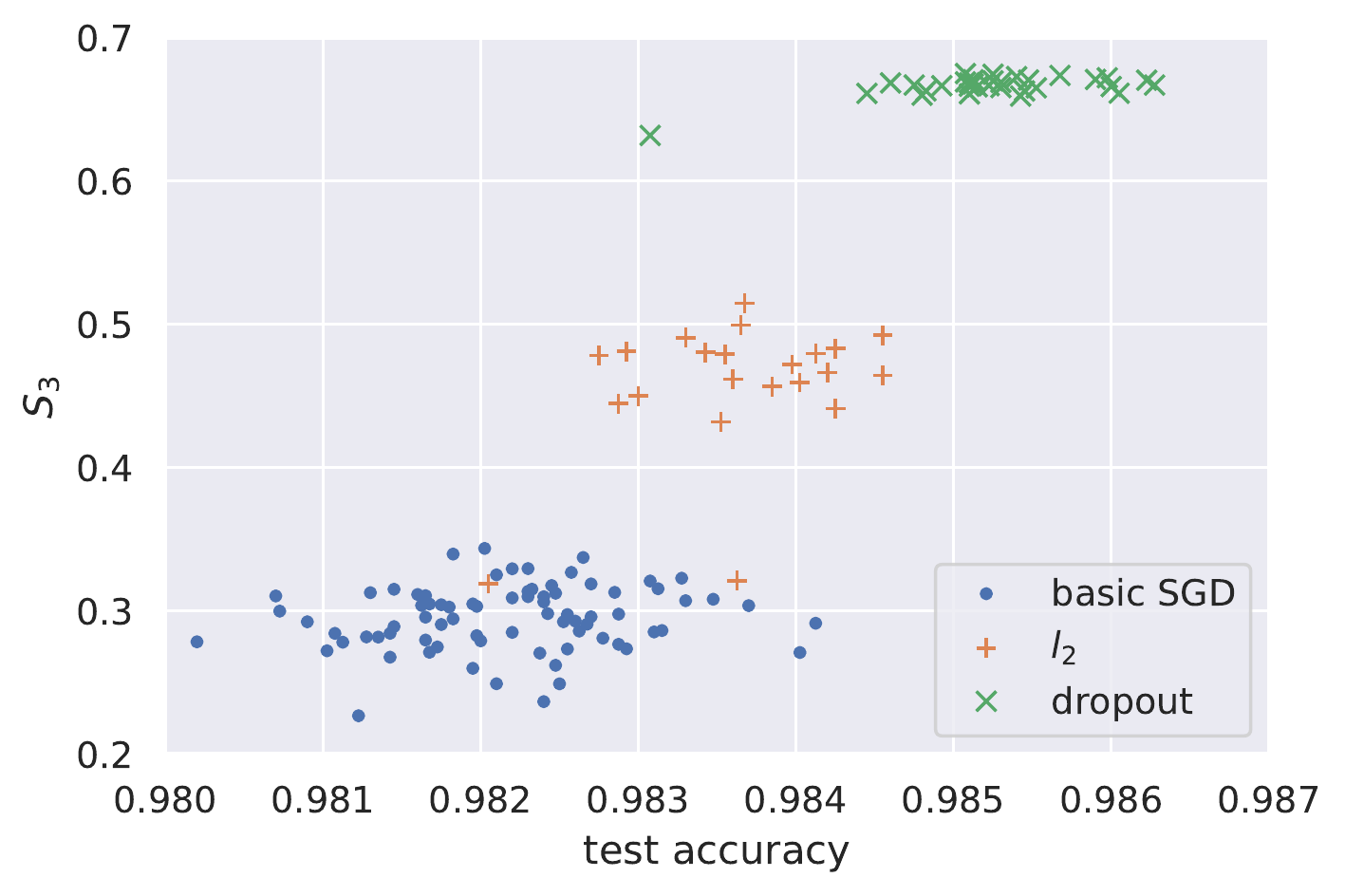}
       \caption{\snr{} for Layer 3}

     \end{subfigure}
     \caption{\snr{} for individual layers, 3-layer DNN applied to MNIST}
     \label{fig:MNIST_simple}
   \end{figure}

Figure \ref{fig:MNIST_simple} shows the results for multiple runs, each using different initial weights and one of the following regularisation schemes: basic SGD (unregularised), L2 regularisation or dropout. For each trained network we calculate $\frac{1}{N}\sum_i^N{S_i}$ (summed over all nodes in the layer) and plot the results against the accuracy over the test dataset. 

From these results we can already see a clear relationship between the accuracy and the \snr{} parameters. For each layer, the more effective regularisation scheme generates the higher \snr{}. Surprisingly, this can be seen even though we are averaging over a large number of nodes in each case.\par

Based on these results, we have some encouragement that generalisation performance is related to how well nodes comply with their respective optimal SNR weight configurations. Before applying this to a wider range of training techniques and networks, there are two observations that suggest a need for refinement of our analysis:
\begin{itemize}
    \item{Even in this simple example, we see that the different training schemes do not improve the \snr{} of each layer in the same way.  So, for example, regularisation A might optimise better than regularisation B on layer $n$, but worse on layer $n-1$. While the differences are not significant for this example, in more complex cases it may not be clear how to judge between the two.}
    \item{As defined, \snr{} is a measure of deviation from the optimal condition. This is not necessarily a useful quantity if we are comparing the performances of different networks across multiple layers, since deviation from optimality does not translate directly to raw performance. That is, a large deviation from a good optimum is not necessarily worse than a small deviation from a poor optimum. Again, this makes comparison difficult.}
\end{itemize}

\section{\snrgain{}}

The above suggests the importance of \snr{} for a simple case. But for more general cases, we would like to have a quantitative measure that is easier to compare across different cases and, ideally, one that can characterise the SNRs of whole network. \par

As noted in the previous section, \snr{} measures deviation from optimal SNR. To enable more direct comparisons, it would be better to have a value that measures how much a given node improves the SNR compared to a minimal performance alternative.  This is analogous to the `SNR gain' used in antenna engineering, which compares the SNR at the output of a phased array with that of a reference antenna (see, for example, \cite{warnick_maaskant_ivashina_davidson_jeffs_2018}).  We will borrow the terminology.\par

To be a suitable reference for comparison, the `minimal performance alternative' must provide an output close to the `signal' that emerged from the training process. With this in mind, retain our assumption that $\sum w_{ij} x_j$ can be treated as the `signal' and look for the input that is best correlated with this post-training output of the node. We then calculate a second \snr{} value, $S_i'$, using weights that are non-zero only for this input. That is, we fix $c_j$ and then use $w_{ij}'$ defined by
\begin{equation}
    w_{ij}' =  \begin{cases}
                \begin{aligned}
                & \sign(c_k) && \text{if }  j=\text{argmax}_k \left(\vert c_k \vert \right)   \\
               & 0 && \text{otherwise}\\
               \end{aligned}
            \end{cases}
\end{equation}

For this choice,
\begin{equation} 
S_i' = \frac{ \sum _{j}^{}w_{ij}'c_{j}}{ \| \vec{w'_i} \|  \| \vec{c} \| }  =  \frac{\max_{i} \left( c_{i} \right) }{ \| \vec{c} \| } 
\end{equation}
We then estimate the \emph{\snrgain{}} by calculating the ratio of the $S_i$ and $S_i'$, giving
\begin{equation}
G_i=\frac{S_i}{S_i'} =\frac{ \sum _{j}^{}w_{ij}c_{j}}{ \| \vec{w_i} \| \max_{j} \left( \vert c_{j} \vert \right) } 
\label{eq:F}
\end{equation}

As with \snr{}, we can average the \snrgain{} to give a expression for a layer in the network. We would like to go further and combine them to give an overall figure-of-merit for the entire network. In this paper, we use the following

\begin{equation}
    G^{(m..n)} = \sum_{j = \text{layer~m}}^{\text{layer~n}}{G_j} \label{eq:netF}
\end{equation}
Two comments should be made about \eqref{eq:F} and \eqref{eq:netF}: 
First, we have to be careful when applying $G_j$ to softmax layers, since such layers have an ambiguity with respect to our figures-of-merit and prove to be more variable (see appendices \ref{sec.softmax} and \ref{sec.limitations}).
Second, we are not claiming that $G^{(m..n)}$ is generally applicable. It will be seen that it is a valuable expression for the networks discussed here, but it has been selected empirically and does not take into account some subtleties of more general network weight sets. Further work is required to extend this metric to a more general expression.\par

\section{\snrgain{} applied to MNIST, CIFAR-10}

Consider the \snrgain{} expressions applied to 2 cases: MNIST with a 3-layer fully connected network and CIFAR-10 \cite{CIFAR} with a 5-layer CNN. In both cases, we used a number of different training approaches (regularisation, data augmentation, etc.) to ensure a good spread of performances. For each network we calculate $G_j$ for each layer and $G^{(1..n-1)}$. (Due to the \emph{softmax} observation above, we find that looking at  $G^{(1..n-1)}$ and $G_n$ separately is more instructive than $G^{(1..n)}$.)\par

\begin{figure}[!htb]
     \begin{subfigure}[b]{0.3\linewidth}
       \includegraphics[width=\linewidth]{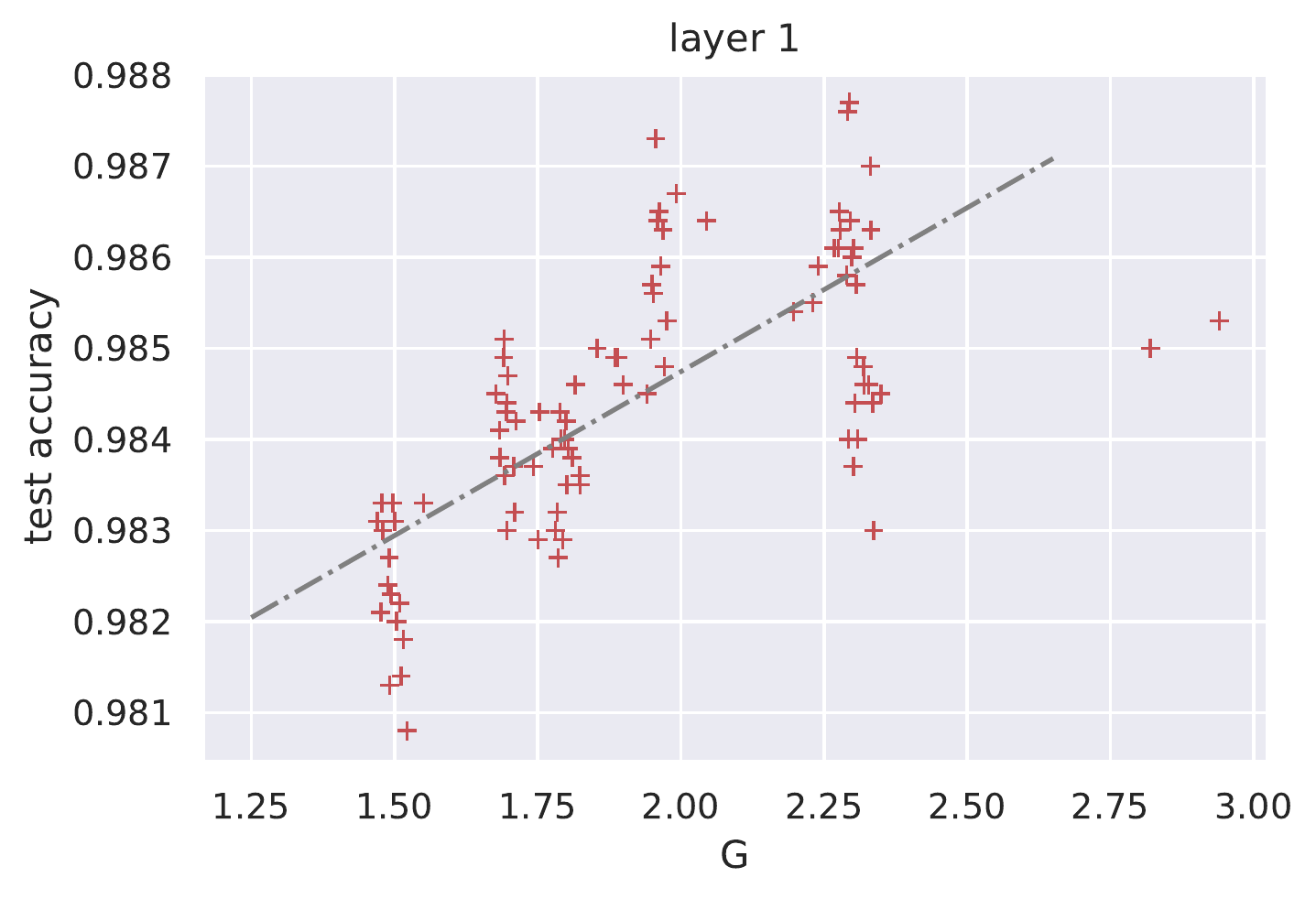}
       \caption{$G$-parameter for Layer 1}

     \end{subfigure}
      \hfill
     \begin{subfigure}[b]{0.3\linewidth}
       \includegraphics[width=\linewidth]{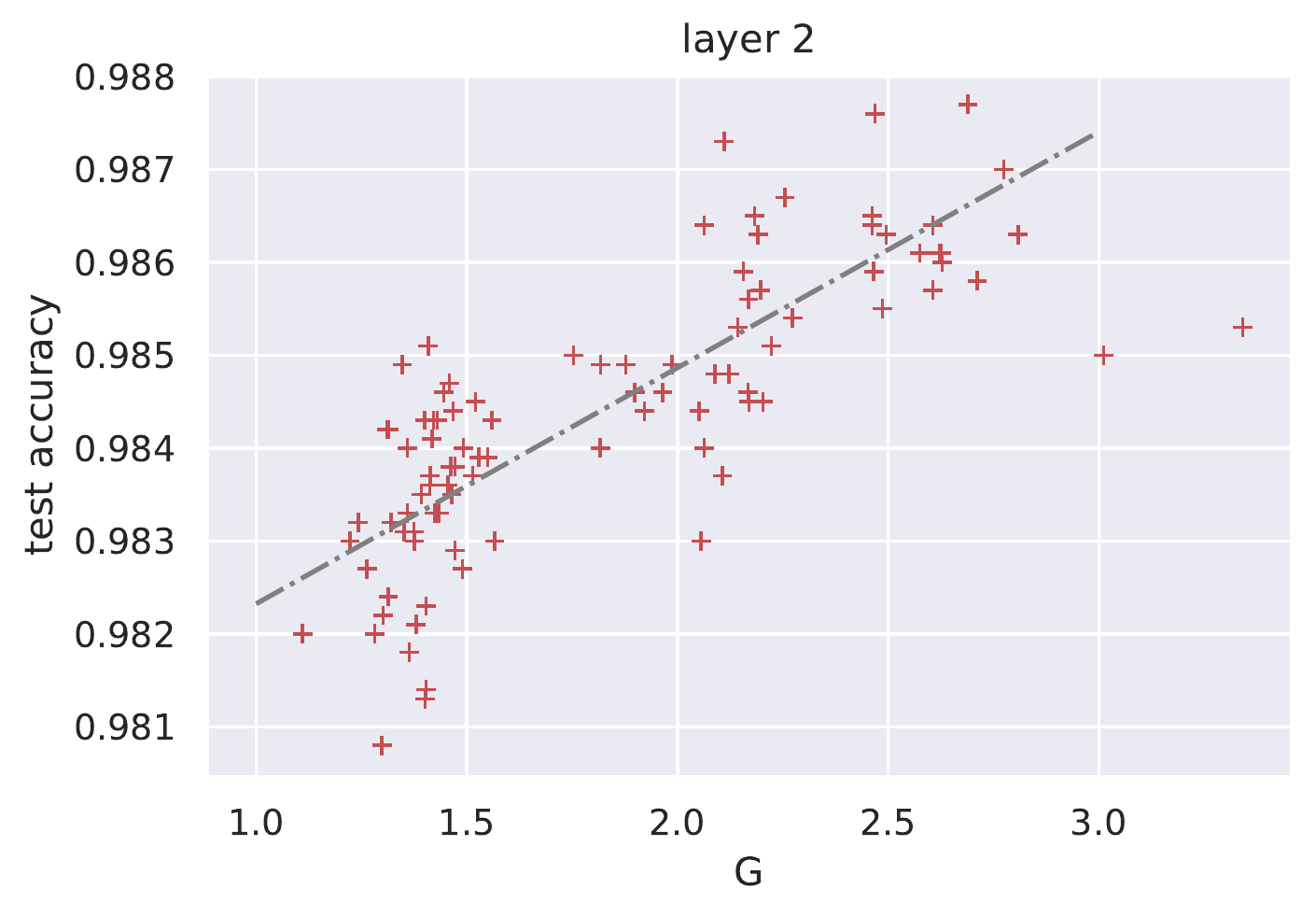}
       \caption{$G$-parameter for Layer 2}
     \end{subfigure}
   \hfill
     \begin{subfigure}[b]{0.3\linewidth}
       \includegraphics[width=\linewidth]{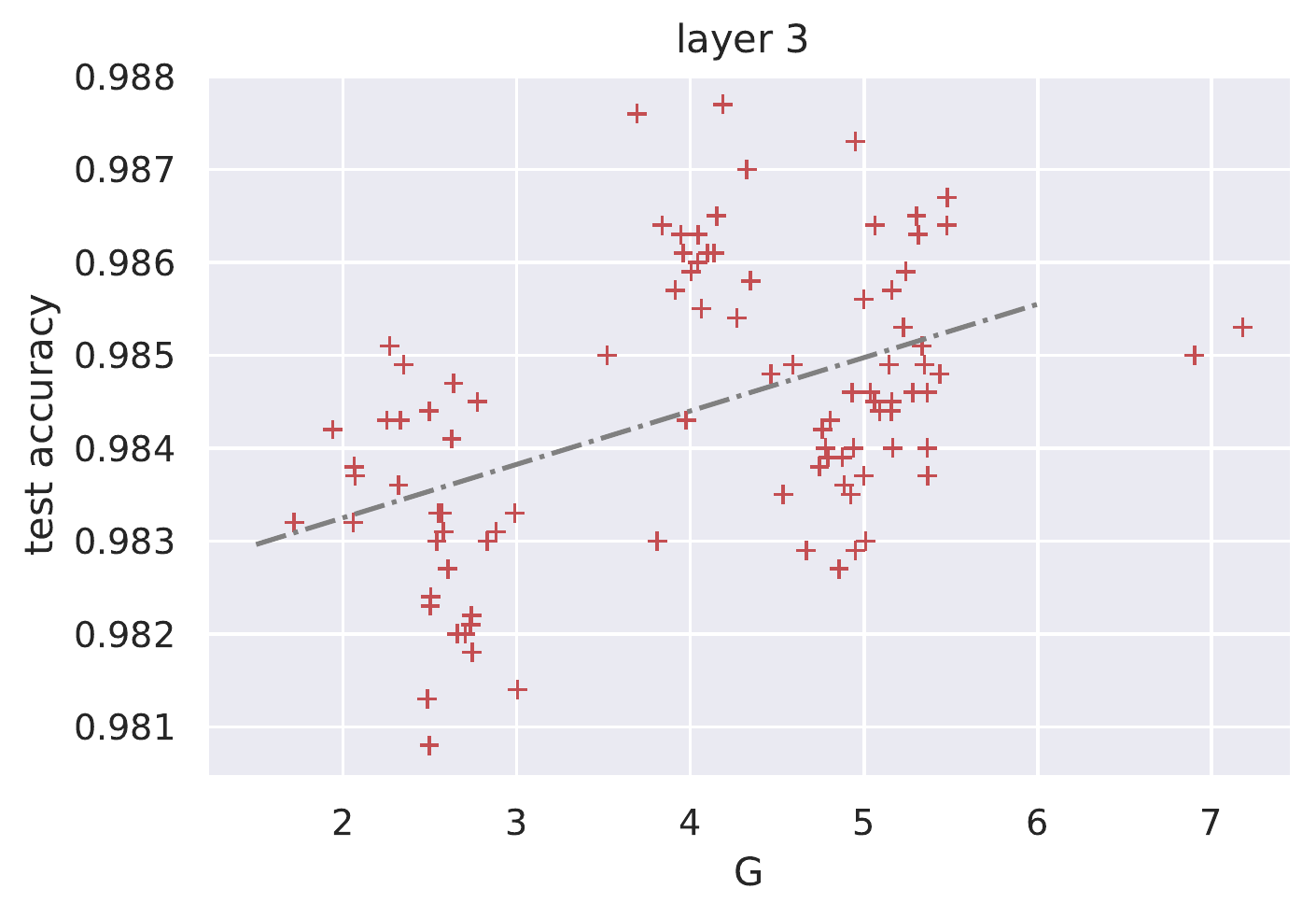}
       \caption{$G$-parameter for Layer 3}

     \end{subfigure}
     \caption{$G$-parameters for individual layers, 3-layer DNN applied to MNIST}
     \label{fig:MNIST_3L}
   \par\bigskip
     \centering
     \begin{subfigure}[b]{0.4\linewidth}
       \includegraphics[width=\linewidth]{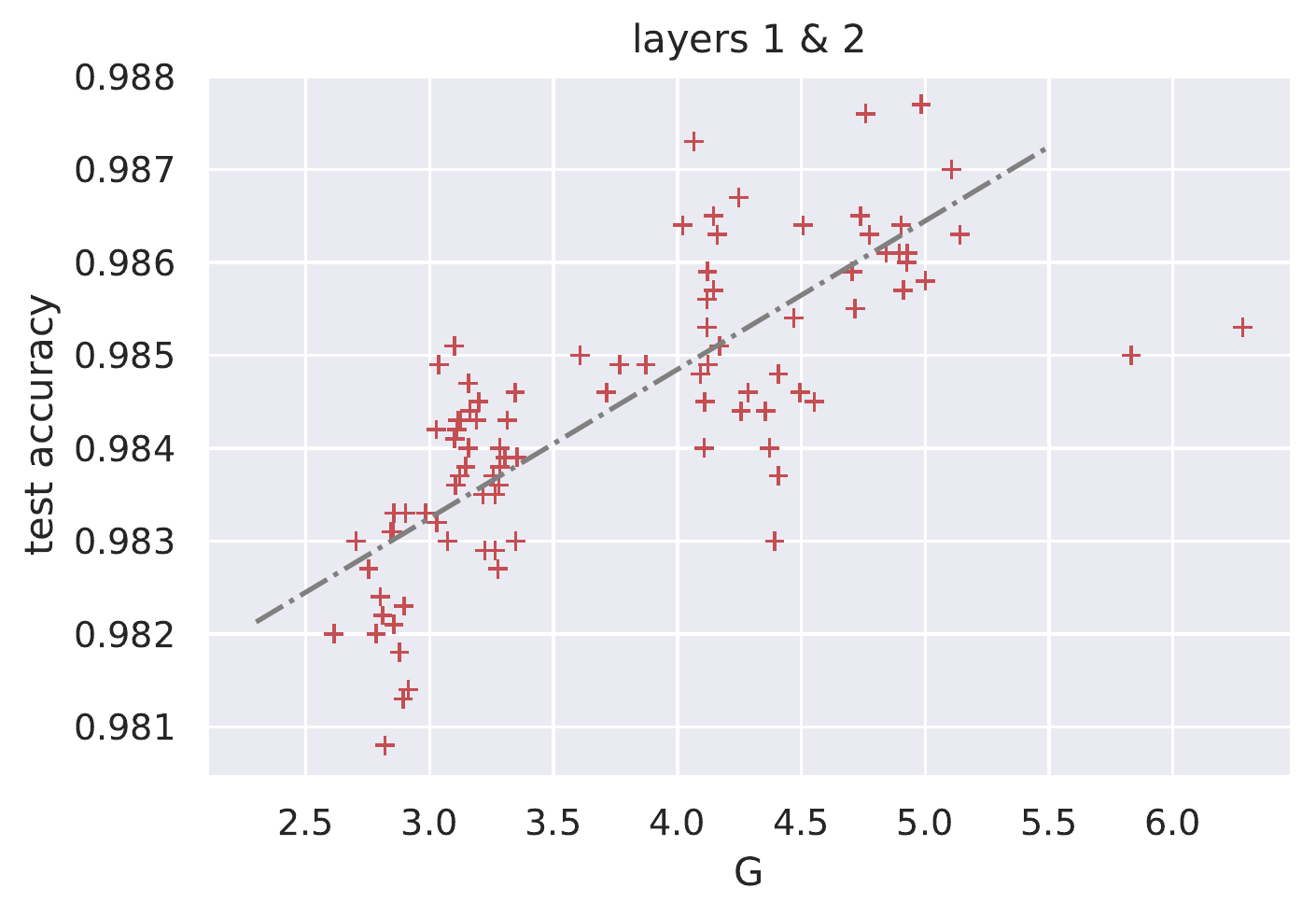}

     \end{subfigure}
     \caption{$G^{(1..2)}$-parameter, 3-layer DNN applied to MNIST}
     \label{fig:MNIST_3L_tot}
     
     \par\bigskip
    \captionof{table}{Statistical coefficients for $G$-parameters vs. test accuracy\\ (3-layer DNN applied to MNIST)}
    \label{tab:mnist}
    \centering
    \begin{threeparttable}
    \begin{tabular}[t]{llccccccc}
    \hline
    &&$G_1$&$G_2$&$G_3$&&$G^{(1..2)}$\\
    \hline
    &$r^2$& 0.501& 0.666& 0.201&&0.641\\
    Spearman &$\rho$& 0.700& 0.818& 0.416&&0.799\\
     &p-value& 0.000& 0.000& 0.000&&0.000\\
    \hline
    \end{tabular}
    \begin{tablenotes}\footnotesize
    \item{Note: $r^2$ calculated with outliers removed}
    \end{tablenotes}
    \end{threeparttable}
    \par\bigskip
   \end{figure}

\begin{figure}[!htb]
     \begin{subfigure}[b]{0.3\linewidth}
       \includegraphics[width=\linewidth]{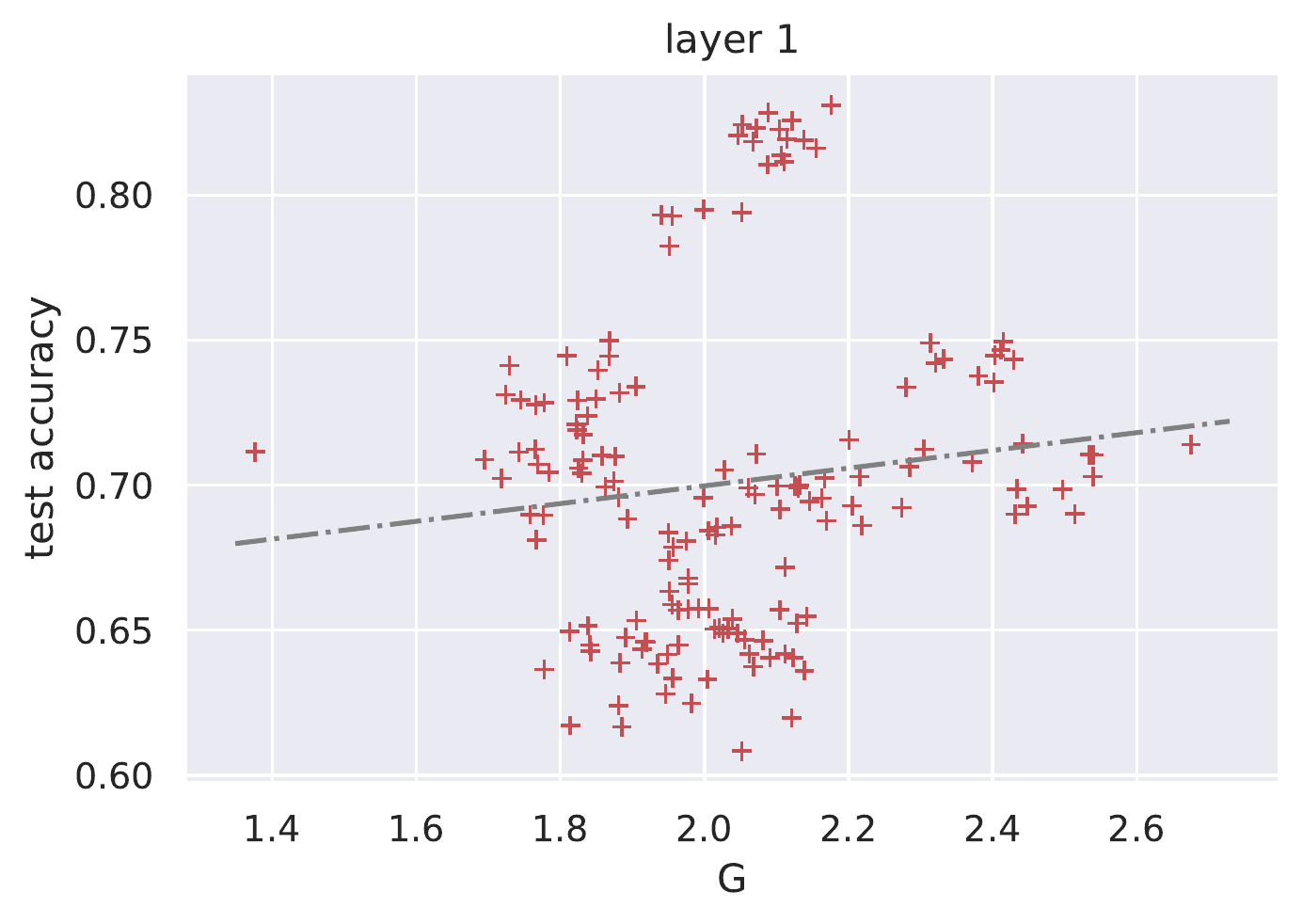}
       \caption{$G$-parameter for Layer 1}
     \end{subfigure}
   \hfill
     \begin{subfigure}[b]{0.3\linewidth}
       \includegraphics[width=\linewidth]{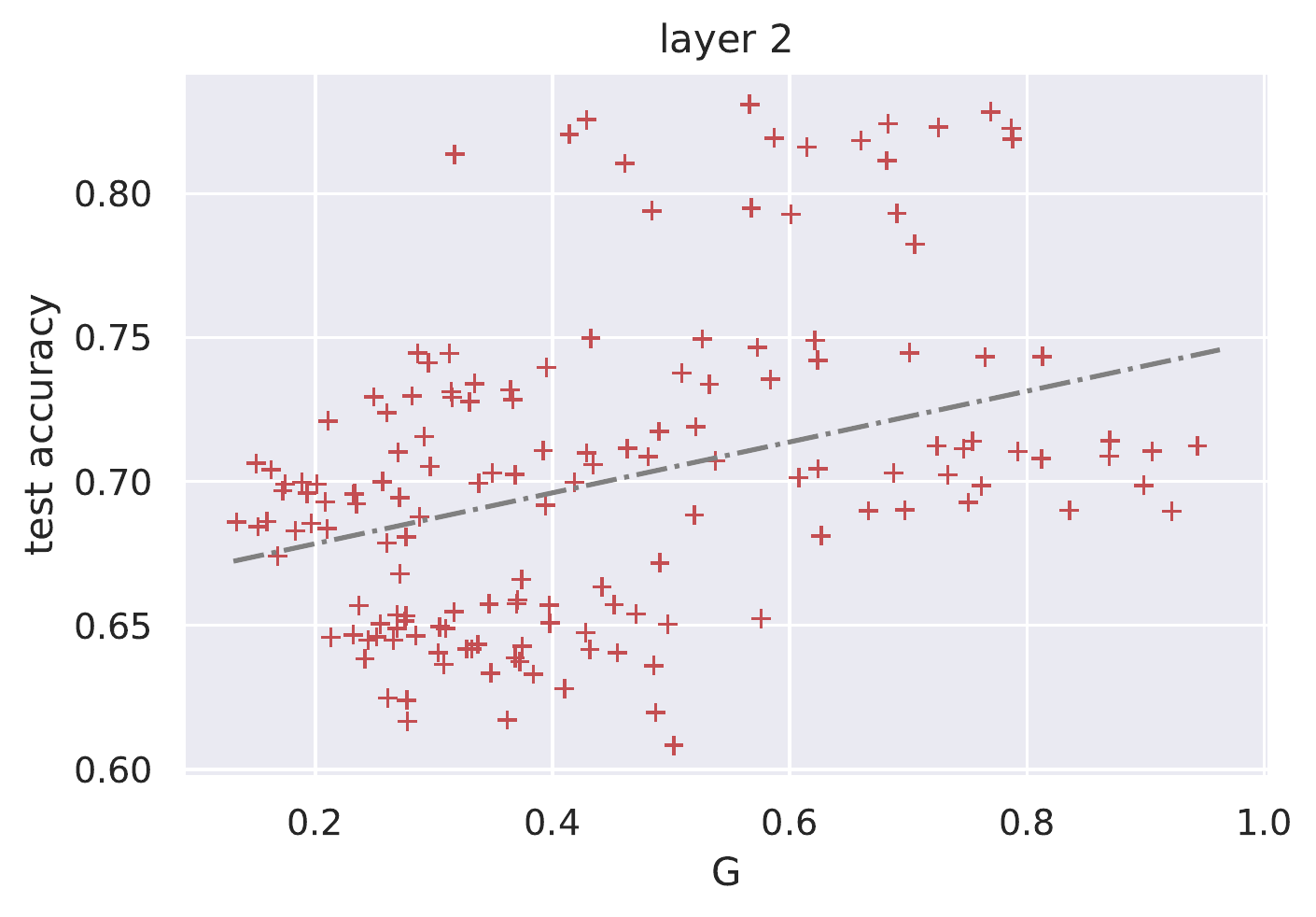}
       \caption{$G$-parameter for Layer 2}
     \end{subfigure}
    \hfill
      \begin{subfigure}[b]{0.3\linewidth}
       \includegraphics[width=\linewidth]{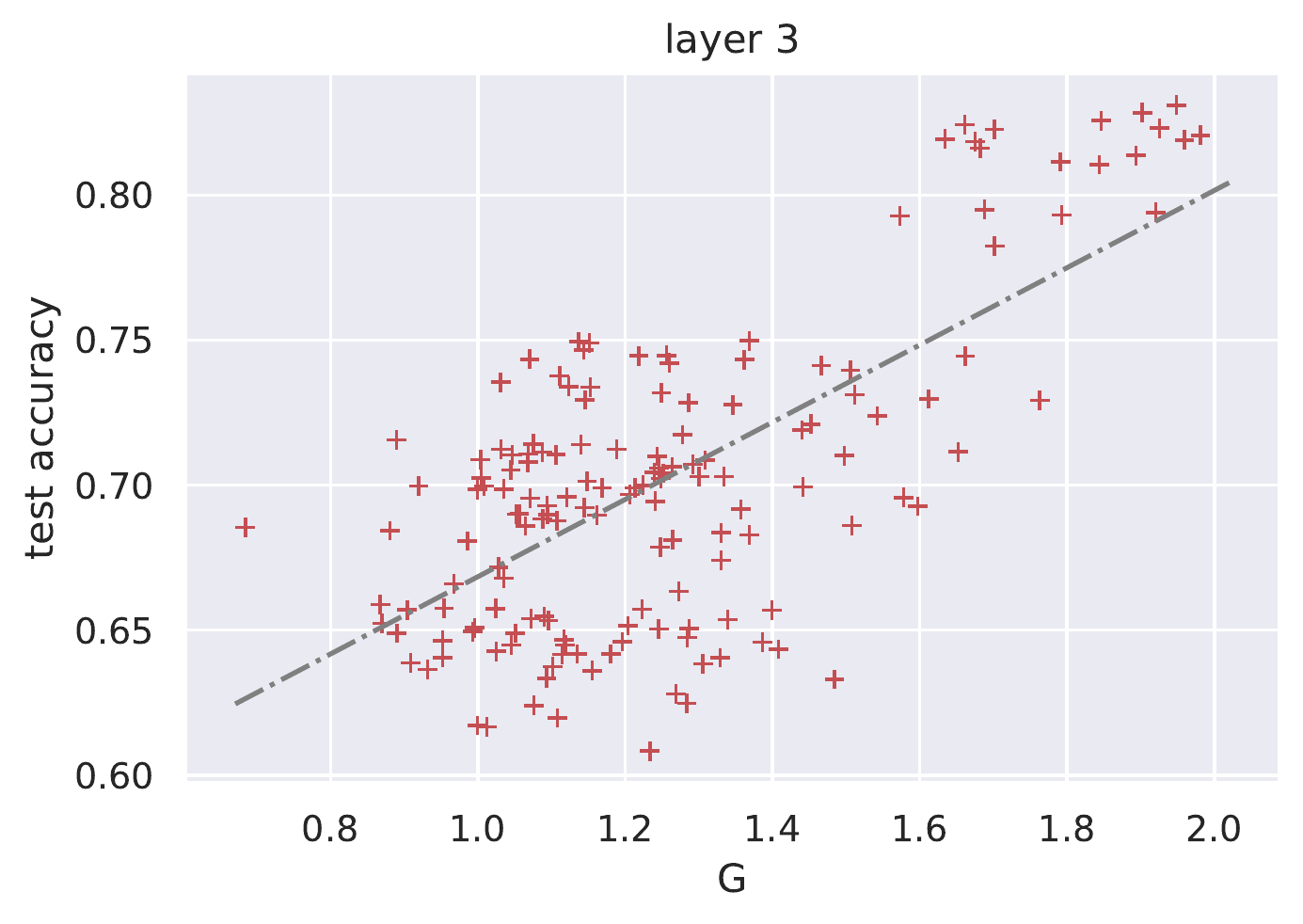}
       \caption{$G$-parameter for Layer 3}
     \end{subfigure}
    
    \centering
     \begin{subfigure}[b]{0.3\linewidth}
      \includegraphics[width=\linewidth]{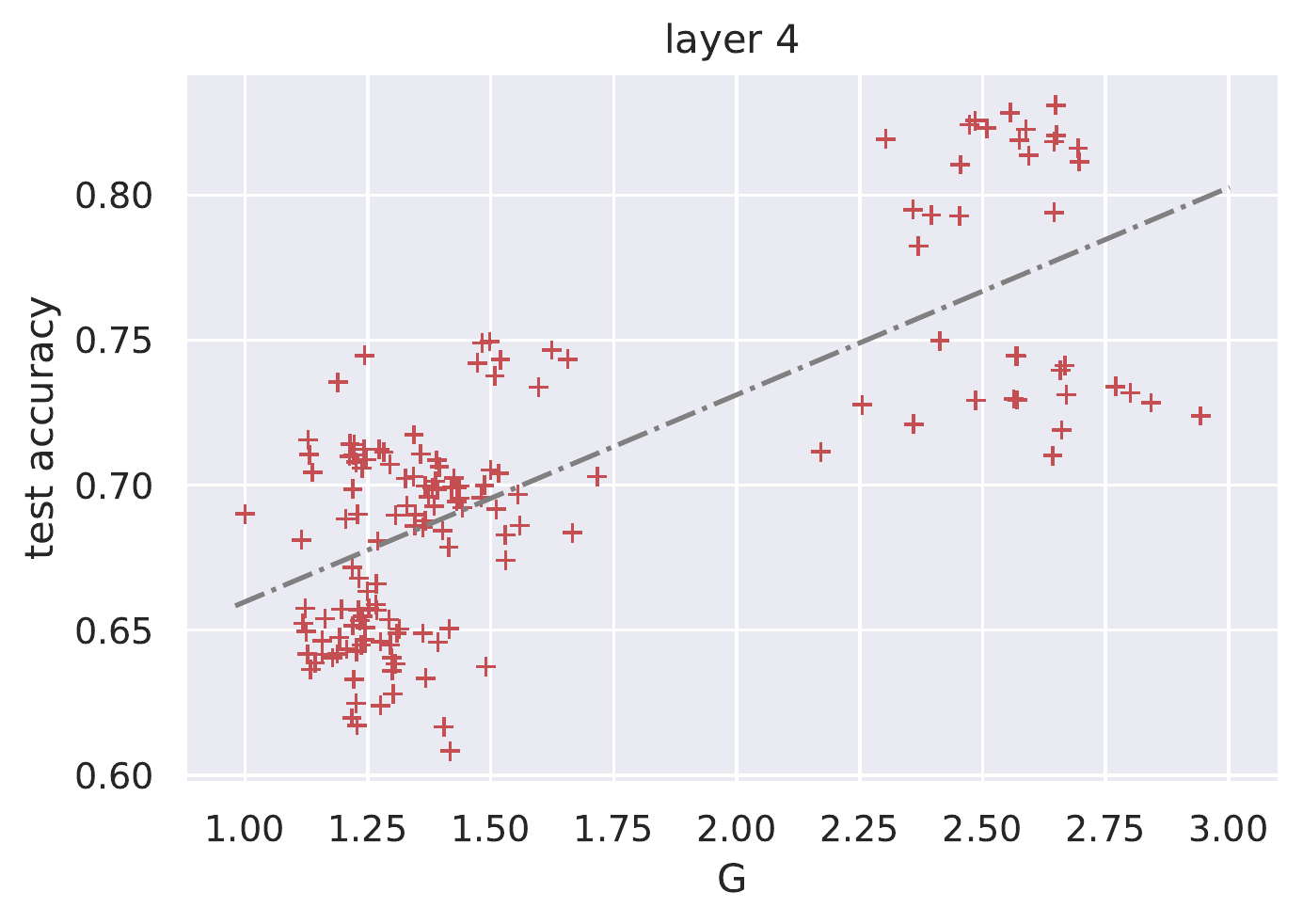}
      \caption{$G$-parameter for Layer 4}
     \end{subfigure}
     \begin{subfigure}[b]{0.3\linewidth}
       \includegraphics[width=\linewidth]{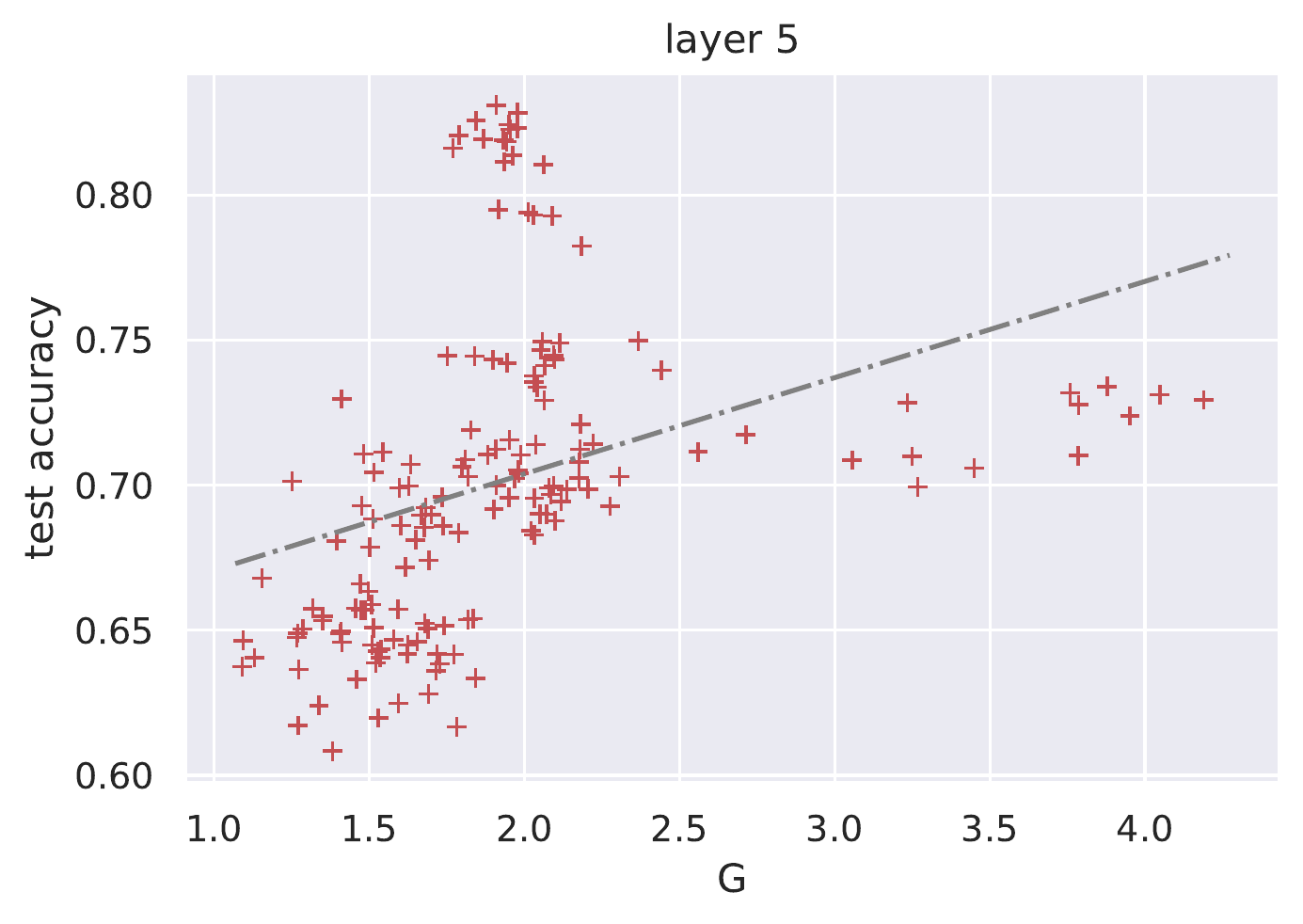}
       \caption{$G$-parameter for Layer 5}
     \end{subfigure}

     \caption{$G$-parameters for individual layers, 5-layer CNN applied to CIFAR}
     \label{fig:CIFAR_5L}
    \par\bigskip
     \centering
     \begin{subfigure}[b]{0.4\linewidth}
       \includegraphics[width=\linewidth]{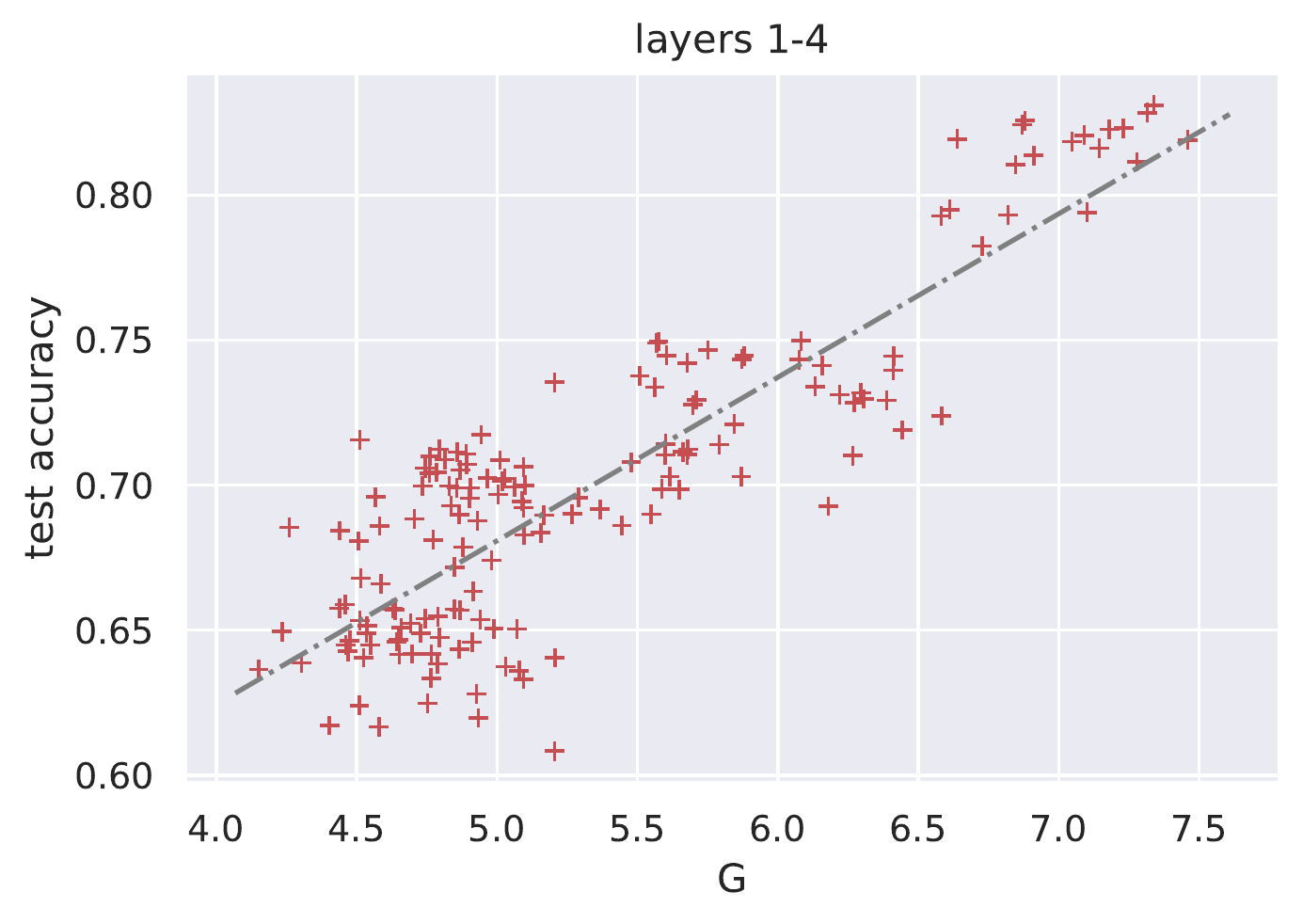}

     \end{subfigure}
     \caption{$G^{(1..4)}$-parameter, 5-layer CNN applied to CIFAR-10}
     \label{fig:CIFAR_5L_tot}
    \par\bigskip
    \captionof{table}{Statistical coefficients for $G$-parameters vs. test accuracy \\ (5-layer CNN applied to CIFAR-10)}
    \label{tab:cifar}
    \centering
    \begin{tabular}[t]{llccccccc}
    \hline
    &&$G_1$&$G_2$&$G_3$&$G_4$&$G_5$&&$G^{(1..4)}$\\
    \hline
    &$r^2$&0.019&0.137  &0.465&0.545&0.132&&0.750\\
    Spearman &$\rho$&0.143&0.394  &0.480&0.601&0.616&&0.790\\
     &p-value&0.065&0.000&0.000&0.000&0.000&&0.000\\
    \hline
    \end{tabular}
    \par\bigskip
   \end{figure}

The results are given in figures \ref{fig:MNIST_3L}, \ref{fig:MNIST_3L_tot}, \ref{fig:CIFAR_5L} and \ref{fig:CIFAR_5L_tot}, with each plotting test accuracy against the $G$-parameters, plus best-fit lines. Tables \ref{tab:mnist} and \ref{tab:cifar} give the $r^2$ and Spearman coefficients between $G$ parameters and the test accuracy.

For the MNIST case, we see good correlations for the $G_1$, $G_2$ and $G^{(1..2)}$ parameters\footnote{Note that there are two outliers, which have high $G$ values. These are the result of training for an extended period with $l_2$ regularisation applied. This is a common feature. It is associated with a very high activation rate ($>0.9$) on the layer 2 output, so we conjecture that it because -- from an \snr{} point-of-view -- the distinction between layer 2 \& layer 3 breaks down.}. For the CIFAR-10 case, we see a more surprising result: while the individual layers show a weak relationship with test accuracy, $G^{(1..4)}$ has a very clear correlation. In both cases, the correlation between test accuracy and $G$ of the final layer is weaker. As already noted, there is more variability in the final layers, partly due to subtleties of how the softmax ambiguity translates into actual SNR performance (appendix \ref{sec.softmax}).\par

\section{Discussion}
Based on the above results, there is good evidence that \snr{} is correlated with DNN generalisation performance. Given the intuition that reduction of noise will improve classification, it seems reasonable to conclude that \snr{} is a contributing factor to the accuracy.  As noted previously, this in turn promotes the idea of a close relationship between \infomax{} and generalisation.\par

By looking at the expression for optimal weights and the figures-of-merit, it is relatively straightforward to identify the properties of a weight set that will optimise SNR: If there are correlated inputs to a node, they should have comparable weighting to give the best noise performance\footnote{Note that expressing it this way hints at the relationship between our discussion and the ability of a DNN to find features in underlying data. In visual problems, human identifiable features are precisely pixels that are correlated, that is, are frequently active at the same time. So, we are effectively demonstrating that the ability to find visual features is related to generalisability.}. If we consider that training has led to an appropriate `signal' from a node, then this weighting should be high if they are correlated with the `signal' and low otherwise.\par 

The results also indicate that \snr{} is a consequence of the way the training is implemented -- we have seen empirically that different training approaches generate nodes and networks that are more or less sensitive to noise. With an idea of what makes weight sets successful from an SNR perspective, we now consider broader training questions: First, how this relates to over-fitting and, second, why regularisation techniques might enhance \snr{}. \par

As a preliminary for this discussion and given the above observations, it is useful to have in mind the shape of the error function when there are correlated inputs. Consider an ideal case, where a subset of inputs are perfectly correlated and, hence, interchangeable. In the absence of noise, any linear combination of this subset will contribute identically to the output; the error function minima will consist of a region rather than a point and all points in the region will effectively be equivalent, providing the same results.\par

From the analysis in section \ref{sec.groundwork}, when noise is introduced the points on the region become distinguished since some points will give better performance than others. \par

Successful training will, in general, find a point on this regional minima. The point may or may not be optimal for SNR. Whether this point \emph{will} be optimal will depend on the exact conditions of the training. Equivalently, the results above imply that generalisation performance will depend on where in the region the parameters are when training stops. (We note the relationship between this point and the learning dynamics discussion in \cite{ConvergenceTime}. Once the regional minimum is reached it is slow or impossible to move to better points within the region.)\par
 
We can compare this with the frequent assumption that a sub-optimal training result is due to finding a `local minimum'. In this case, there may be no `local minimum'; the problem is that there is a (possibly global) regional minimum that contains some points with higher noise sensitivity than others. \par 

Of course, as we move away from the ideal we may no longer find inputs that are exactly interchangeable, but this is a useful approximation to have in mind\footnote{We briefly note that the discussion here may shed some light on the debate over the relationship between flat/sharp minima and generalisation performance (\cite{Sepp_Hochreiter_Jrgen_Schmidhuber1997},  \cite{Nitish_Shirish_Keskar_et_al_2016},  \cite{Laurent_Dinh_et_al_2017}, \cite{ExploringGeneralization}, \cite{2017arXiv171005468K}).\par
A consideration of \snr{} suggests that there will be a relationship between good generalisation and flat minima. We observe that if the nodes on the first layer of a network have a high \snr{}, then the networks will have regions where the minima are flat (or nearly flat) even if there is a large noise component from other inputs. For low \snr{} in these nodes, this is no longer guaranteed. This suggests that flat minima and good generalisation will frequently be seen together even if the causal connection is not direct.\par The argument here also implies that finding the \emph{right location} on a flat minimum is an important part of the process.}. \par

\subsection{Relationship to over-fitting}

We can clarify this further by comparing \snr{} to over-fitting.  Although over-fitting and \snr{} are both concerned with noise in the data, it is useful to recognise that they are distinct features. Over-fitting is a feature of the training process and the unwanted fitting to the noise in the training data; in contrast, \snr{} is a feature of the network after training is complete and is concerned with reduced sensitivity to noise.\par

To illustrate this, consider a hypothetical scenario where the training data for a DNN has no noise, but the test data is noisy. In this case, there is no possibility of over-fitting the network to the training data, at least in the sense it is normally characterised. However, given the discussion here, we see that it is very likely that some weight sets will perform better on post-training data than others, depending on their sensitivity to the test data noise.\par

In fact, the implicit problem in this idealised example is that the training data is unrepresentative because it is \textit{insufficiently} noisy compared to the overall domain. In effect, we have over-fit to the \emph{lack} of noise.\footnote{We note that a similar effect will be seen if the training batch is sufficiently large to attenuate random noise contributions to weight updates. This suggests an SNR-based reason for small batch sizes being beneficial. cf \cite{2017arXiv171006451S, Nitish_Shirish_Keskar_et_al_2016, EfficientBackprop}}\par

\subsection{Relationship to regularisation methods}
The results above already suggests a more complex relationship between generalisation and regularisation than is generally assumed. We have seen empirical indications that regularisation techniques act to improve the \snr{} of a network. Here we strengthen the connection by outlining the mechanisms that link some common regularisation schemes to improved SNR performance.\par

We begin by observing that straightforward SGD provides no guarantees of finding an SNR-optimal solution. Indeed, we expect that it will frequently converge to a solution that is not \snr{}. We observe that weight updates generated by SGD are highly dependent on other weights in the network and not just on the properties of the input samples themselves. So, two inputs may have correlated samples, but have very different weight updates due to the attenuation or enhancement provided by the rest of the network. In fact, there will often be a `success breeds success' aspect, where strongly weighted inputs of a node are enhanced, even while noise sensitivity would be improved if some of the weakly inputs were also emphasised.\par

Referring back to the previous section, if we consider the training minima to be a region, SGD will frequently move the model to a point with poor \snr{}. In contrast, a good regularisation technique will compensate for this and move the solution towards points with high \snr{}. \par

The simplest example is $l_2$ regularisation. Recall that $l_2$ regularisation adds a term $\sum{w_{ij}^2}$ to the loss function. The term is typically used as a mechanism ensuring there is no over-fitting, but there is also a very specific advantage for \snr{}. Goodfellow \emph{et al} \cite{Ian_Goodfellow__et_al_2016} note that $l_2$ regularisation "shrink[s] the weights on features whose covariance with the output target is low..." This alone will improve SNR performance. However, from the discussion here, there is another equally important aspect: it will act to equalise the weights on inputs that are correlated with the output of a node. By attempting to minimise $\sum{w_{ij}^2}$, we encourage the training process to use as many correlated inputs as it can.\par

Dropout \cite{Srivastava_dropout, Geoffrey_E._Hinton_et_al_2012} has a similar effect in balancing weights on correlated inputs. Recall that dropout randomly removes nodes (and associated links) during the training process. When inputs to a node is removed, the weight update will favour other inputs that provide the same information as those that are missing. If a strongly weighted input is removed, other inputs that are correlated, but have weaker weighting, will be enhanced. Conversely, due to the re-scaling used in the dropout algorithm, when weaker links are removed, dominating links will be de-emphasised. Over a number of iterations, weights on inputs carrying correlated information will tend to balance and the \snr{} will increase. \par     

Regularisation via the addition of noise to training data can be viewed in multiple ways relevant to this discussion. First, Bishop \cite{L2Noise} has shown that adding noise is equivalent to $l_2$ regularisation, so the above discussion can be carried across. Alternatively, we can consider that during training, added noise weakens the `success breeds success' aspect of SGD; due to noise, a useful but under-emphasised path will occasionally have a larger weight update than it might otherwise; as a result the weights of correlated inputs will get closer. Looking at it more abstractly, and referring back to the hypothetical scenario above, we noted that noise-free training data could lead to a potential `over-fitting to the lack of noise'. In contrast, increasing the noise in the training data will encourage it to be less sensitive to noise in general. (See also \cite{2017arXiv171006451S}.)\par

Without going into details, we also note that Batch Normalisation \cite{Nitish_Shirish_Keskar_et_al_2016} also weakens the `success breeds success' aspect of SGD in a way that will act to equalise the weights of correlated inputs. \par

Although these are by no means rigorous proofs, the ability to identify mechanisms that link regularisation schemes with \snr{} gives weight to our proposal that SNR considerations are related to generalisation performance.

\subsection{Applications}

Although the aim of this paper is to demonstrate \snr{} is an important factor for generalisation performance, we note that this is not simply an abstract discussion; \snr{} has potential as a tool in DNN development. As examples, initial investigations have show that it can be used for the following:
\begin{itemize}
    \item {\textbf{New regularisation approaches:} By understanding the way regularisation affects node-level \snr{}, we can derive alternative regularisation techniques that focus on this aspect. }
    
    \item{\textbf{Selecting regularisation combinations:} We have already noted that different regularisation schemes are more beneficial to some layers in a network than others. By looking from an \snr{} perspective, we can begin to identify how a regularisation scheme functions at the level of individual layers. This opens the possibility of adjusting or combining regularisation schemes to compensate for weaknesses. For example, experiments have shown that we can improve performance of a training scheme by selectively applying dropout to layers that had poor \snr{}.}
    
    \item{\textbf{Node and link thinning:} By looking at the \snr{} and identifying inputs that have weak correlations, we have an alternative metric to guide thinning of a network. We find that this gives us a productive criteria for removing superfluous inputs or nodes with negligible impact on overall performance. }
\end{itemize}

\section{Conclusion}

We have demonstrated that for a trained DNN, there is a correlation between the \snr{} of nodes and the generalisation performance of the network. Networks that are equally well trained for a given the dataset may exhibit better or worse generalisation depending on the detailed weight choices at node-level (even in the absence of over-fitting). This gives added weight to the \infomax{} and the idea that quality of information flow though a network impacts performance. It also emphasises the connection between the ability to extract features (characterised by correlated inputs) and generalisation.\par

Looking at \snr{} provides an alternative perspective on DNN configuration and performance. It opens up ways to discuss the properties of individual layers in a network and gives a different view on the benefits of regularisation and other training enhancements. The results here also have value beyond addressing theoretical questions; we have outlined ways that they can be used to guide DNN training and post-processing.\par

 We suggest that this work goes some way to addressing the observations in \cite{Chiyuan_Zhang_et_al_2016}. As noted there, \emph{[w]hile explicit regularizers like dropout and weight-decay may not be essential for generalization, it is certainly the case that not all models that fit the training data well generalize well.} We have shown that there is a characteristic of DNNs that distinguishes between different models that fit the same training set. This characteristic can be enhanced by common regularisation techniques, but they are not a necessary condition for achieving this; many other factors can also push the network configuration to have good \snr{}.\par

\paragraph{Acknowledgments}
The author is grateful to Chiyuan Zhang for taking time to read earlier versions of this work. This work also benefited from discussions with Satwinder Chana on the importance of $G/T$ in Phased Array Antennas and from the availability of Google Colaboratory.

\printbibliography
\newpage

\appendix

\section{Details on experiment setup}\label{sec.setup}
\subsection{Networks}

In the work here, we used two of the simple networks typically provided as introductory examples to classification (e.g. in the TENSORFLOW \cite{tensorflow2015-whitepaper} documentation). Using these had a number of advantages: First, they are quick to train, so a large number of results could be generated easily. Second, a number of different training approaches can easily be implemented to obtain a good range of generalisation performances. Third, the typical performance is not \emph{too} good, so that the variations are clear. Fourth, the \snr{} calculations have low computational cost and the small number of layers aids interpretation.\par

The networks used are
\begin{itemize}
    \item{MNIST: Three fully connected layers with 1024, 1000 and 10 nodes respectively.}
    \item{CIFAR-10: Two convolutional layers with max pooling, each with 64 5x5 kernals per channel. Three fully connected layers with 1000, 500 and 10 nodes respectively.} 
\end{itemize}
All nodes have \relu{} activation applied, except the those in the last layer, which feed a softmax function.\par

In order to get a good spread of results, the training options included unregularised SGD, $l_2$, dropout and basic test set transformations. Additional regularisation methods inspired by \snr{} considerations were also included but detailed descriptions will be addressed elsewhere. \par

Weight updates were limited to the form $\Delta w \sim \epsilon\frac{\partial f}{\partial w}$, with fixed $\epsilon$  (see also \ref{sec.limitations}).

\section{Further points on \snr{} and \snrgain{}}\label{sec.derivations}

\subsection{ Implementation }\label{sec.implementation}
For readers wishing to replicate the approach described here, we note some details in application.\par
\begin{itemize}
    \item{In deriving the formulae there is an implicit assumption that the inputs are scaled such that the noise has the same variance on each input. It is possible that weighting earlier in the network will lead to the noise of each input being scaled by different amounts. Before calculating the \snr{} we should compensate for this to get a meaningful result.  Strictly, we should estimate the noise levels by calculating the cumulative weightings back to the network inputs and scale the inputs appropriately before calculating the \snr{}. However, this is computationally expensive. We find that, in practice, a short-cut can be taken: we assume that we can simply scale each input according to its maximum value across the whole batch, $x_j~\rightarrow~\frac{x_j}{max_{\text{batch}}( x_j)}$.}
\item{Frequently, the results are more informative if we remove from the SNR calculations inputs that have very low levels over all samples. These do not contribute significantly to the final output of the DNN and the low signal means that the associated weights will see little or no change during training.}
\item{When calculating the mean $G$, we have weighted the result for each node using an estimate of the proportion of time it is active over the test batch. This guards against including spurious contributions from nodes that never used. In the majority of cases the difference between unweighted and weighted means is small.} 
\end{itemize}
\subsection{Limitations}\label{sec.limitations}

In the main text, we noted that $G^{(m..n)}$ should not be considered generally applicable. There are two cases that make this clear: First, we recall that we see some outlying points related to long training times with an strongly-weighted $l_2$ regularisation.  As noted in the main text, it is associated with a very high activation rate (>0.9) on the inputs, so we conjecture that it because the distinction between layers breaks down.\par

In addition, if a node has two sets of mutually exclusive inputs (that is, inputs that are never activated simultaneously) and these are weighted similarly, $G_j$ will be larger than for a node that responds only to one of the sets. However, the underlying SNR performance is the same in both cases. It is likely that this is some of the reason for large variability on the final layers since these will frequently be combining mutually exclusive signals associated with the same labels. There is initial evidence that this type of structure may also arise from weight updates beyond the static $\Delta w \sim \epsilon\frac{\partial f}{\partial w}$ (such as Adam \cite{adam}). This is a subject for future study.\par

\subsection{Generalisation to non-\relu{} activations}
In the above derivations, we assumed \relu{} activations on the inputs. We briefly note here the alternation that must be made for the more general case. \par
We denote the activation functions on layer $(m-1)$ by $g_j$ and we explicitly identify $x_j = s_j + n_j$ to be the samples from layer $(m-1)$ \emph{before} the activation function is applied. So the inputs to layer $m$ will be $g_j(x_j)$. If the noise is sufficiently small then we can expand this as $g_j(x_j) \approx g_j(s_j) + g'_j(s_j)n_j$.\par

The leads to an expression for SNR of the form
\begin{equation}
 SNR^{(m)}_i =\frac{\text{var} \left(\sum_j w_{ij}g_j\left(s_j\right)\right) ^{}}{ \sum_j \left(g_j' \left( s_j  \right)  \right)^2 \text{var} \left( w_{ij}n_j\right)}  \approx \frac{\text{var} \left(\sum_j w_{ij}g_j\left(x_j\right)\right) ^{}}{ \sum_j \left(g_j' \left( x_j  \right)  \right)^2 \text{var} \left( w_{ij}n_j\right)} 
\end{equation}\par
From this starting point an alternative expression for optimal weights can be derived for the specific case.
\section{Softmax layers}\label{sec.softmax}
Softmax layers must be treated carefully in this analysis. Recall that softmax is based on the formula
\begin{equation}
    \sigma(\bm{W}\bm{x})_i = \frac{e^{W_{ik}x_k}}{\sum_{j}e^{W_{jk}x_k}}
\end{equation}
Due to cancellations between factors, this is invariant under transformations of the form
\begin{equation}
    W_{ik} \longrightarrow W_{ik} + c_k
\end{equation}
However, this transformation has a non-trivial impact on the \snr{} figure-of-merit. In particular, large offsets common to all weights will generate an artificially high estimate for \snr{} that is not justified by the actual softmax calculation.\par
In order to minimise the impact of this on our calculations, we explicitly apply the above transformation using the mean values over the first index.
\begin{equation}
    W_{ik} \longrightarrow W_{ik} -  \frac{1}{n}\sum_{i=1}^n{W_{ik}}c_k
\end{equation}
where $n$ is the number of nodes in the softmax layer. This proves to be a partial fix, but further work is required to assess in detail how the interaction between nodes enforced by softmax impacts the SNR behaviour.
\end{document}